\documentclass[conference]{IEEEtran}
\IEEEoverridecommandlockouts
\usepackage{cite}
\usepackage{amsmath,amssymb,amsfonts}
\usepackage{algorithmic}
\usepackage[dvipdfmx]{graphicx}
\usepackage{textcomp}
\usepackage{xcolor}
\def\BibTeX{{\rm B\kern-.05em{\sc i\kern-.025em b}\kern-.08em
    T\kern-.1667em\lower.7ex\hbox{E}\kern-.125emX}}
\begin{document}

\title{An Accurate Graph Generative Model

with Tunable Features\\
\thanks{This work was partly supported by JSPS KAKENHI JP23H03379.}
}


\author{\IEEEauthorblockN{1\textsuperscript{st} Takahiro Yokoyama}
\IEEEauthorblockA{\textit{Graduate School of Engineering} \\
\textit{Nagaoka University of Technology}\\
Nagaoka, Niigata, Japan \\
s203193@stn.nagaokaut.ac.jp \\ \\ }

\IEEEauthorblockN{3\textsuperscript{rd} Sho Tsugawa}
\IEEEauthorblockA{\textit{Faculity of Engineering, Information and Systems} \\
\textit{University of Tsukuba}\\  
Tsukuba, Ibaraki, Japan \\
s-tugawa@cs.tsukuba.ac.jp
\vspace*{-.5em}
}
\and
\IEEEauthorblockN{2\textsuperscript{nd} Yoshiki Sato}
\IEEEauthorblockA{\textit{Graduate School of Engineering} \\
\textit{Nagaoka University of Technology}\\
Nagaoka, Niigata, Japan \\
s171039@stn.nagaokaut.ac.jp \\ \\}

\IEEEauthorblockN{4\textsuperscript{th} Kohei Watabe}
\IEEEauthorblockA{\textit{Graduate School of Engineering} \\
\textit{Nagaoka University of Technology}\\
Nagaoka, Niigata, Japan \\
k\_watabe@vos.nagaokaut.ac.jp
\vspace*{-.5em}
}
}

\maketitle

\begin{abstract}
    A graph is a very common and powerful data structure used for modeling communication and social networks.
    Models that generate graphs with arbitrary features are important basic technologies in repeated simulations of networks and prediction of topology changes.
    Although existing generative models for graphs are useful for providing graphs similar to real-world graphs, graph generation models with tunable features have been less explored in the field.
    Previously, we have proposed GraphTune, a generative model for graphs that continuously tune specific graph features of generated graphs while maintaining most of the features of a given graph dataset. 
    However, the tuning accuracy of graph features in GraphTune has not been sufficient for practical applications. 
    In this paper, we propose a method to improve the accuracy of GraphTune by adding a new mechanism to feed back errors of graph features of generated graphs and by training them alternately and independently.
    Experiments on a real-world graph dataset showed that the features in the generated graphs are accurately tuned compared with conventional models. 
\end{abstract}

\begin{IEEEkeywords}
Graph generation, Conditional VAE, LSTM, Graph feature, Generative model.
\end{IEEEkeywords}

\section{Introduction}

In fields regarding communication networks, graph generative models have a wide variety of applications such as network synthesis for simulations, emulation of information spreading on networks, link prediction on social networks, etc.
Generative models for graphs can be categorized into two types: stochastic and
learning-based models.
Stochastic models focus on reproducing only a single-aspect feature of graphs (e.g., scale-free feature). 
On the other hand, learning-based models aim to learn features directly from a graph dataset and reproduce graphs that have similar features to the graph dataset, thus reproducing features in every single aspect.
We have proposed a generative model for graphs, GraphTune~\cite{b1}, that allows continuous tuning of specific features while maintaining the reproducibility of the other graph features. 
Although GraphTune has succeeded in making features of a generated graph change depending on user-specified values, there still remains an issue on the tuning accuracy of features.


In this paper, we propose to extend GraphTune by adding a \textit{feature estimator} that feeds back information on features of graphs reconstructed by GraphTune.
These two models, the Long Short-Term Memory (LSTM)-based feature estimator and GraphTune, are trained independently of each other by an \textit{alternate training algorithm} to avoid target leakage in feature.
GraphTune with the feature estimator enables tuning specific features more accurately than GraphTune while keeping the reproducibility of GraphTune in every single feature.
\section{GraphTune}

GraphTune treats graph data as a graph sequence converted by a DFS code and learns to reconstruct an input sequence in the Conditional Variational AutoEncoder (CVAE) framework.
DFS code converts graphs into sequences of unique edges through depth-first search.
In CVAE, the sequence is processed by an encoder, and then the decoder generates a sequence. 
Lastly, the generated sequence is converted to a generated graph. 
A condition vector is input to the encoder and the decoder to tune a specific feature.
The elements of the condition vector mean the specified value of features focused on (e.g. values of average shortest path length, cluster coefficient, the power-law exponent of the degree distribution, etc.). 


\subsubsection*{Encoder}
  Encoder learns to map from sequence data and condition vector to a multivariate normal distribution
according to graph features. 
A latent vector is sampled from that mapped distribution.

\subsubsection*{Decoder}
  Decoder learns to reconstruct an input sequences from the latent and the condition vectors.
  The first input is generated with the vectors, and a subsequent sequence is recursively generated by LSTM.
  When we generate a graph using GraphTune, we input a sampled latent vector and a specified condition vector to the decoder and obtain a sequence representing a feature-tuned graph. 


\section{Proposed Model}

\begin{figure}[tb]
  \centering
  \includegraphics[keepaspectratio, scale=0.45]{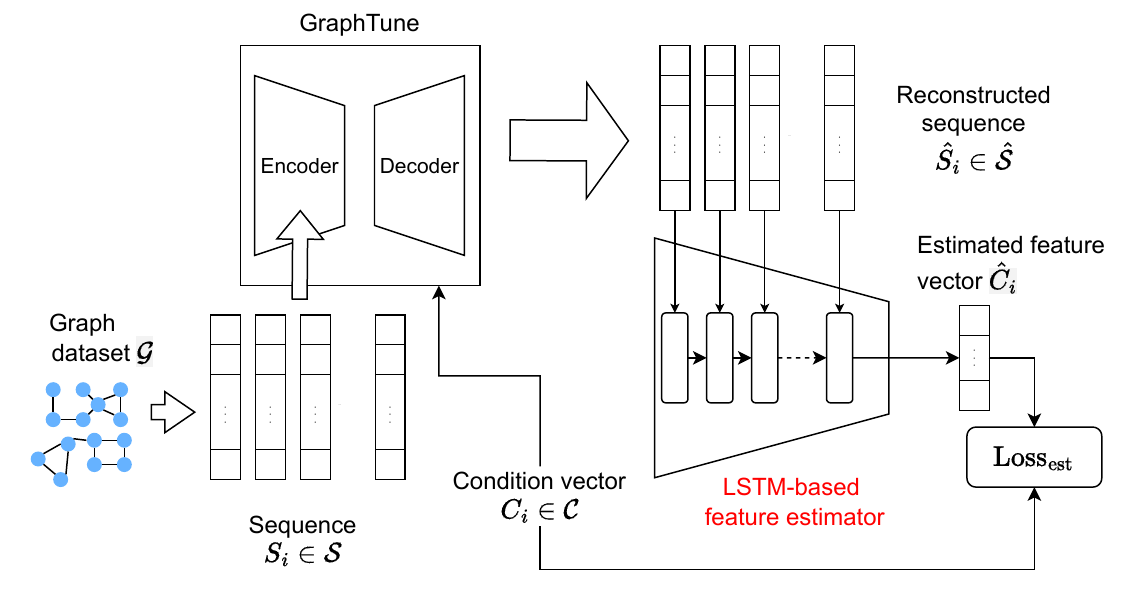}
  \vspace*{-1em}
  \caption{Proposed model composed of GraphTune and a feature estimator.}
  \label{fig:proposed_model}
\end{figure}

\begin{figure}[tb]
  \centering
  \includegraphics[keepaspectratio, scale=0.19]{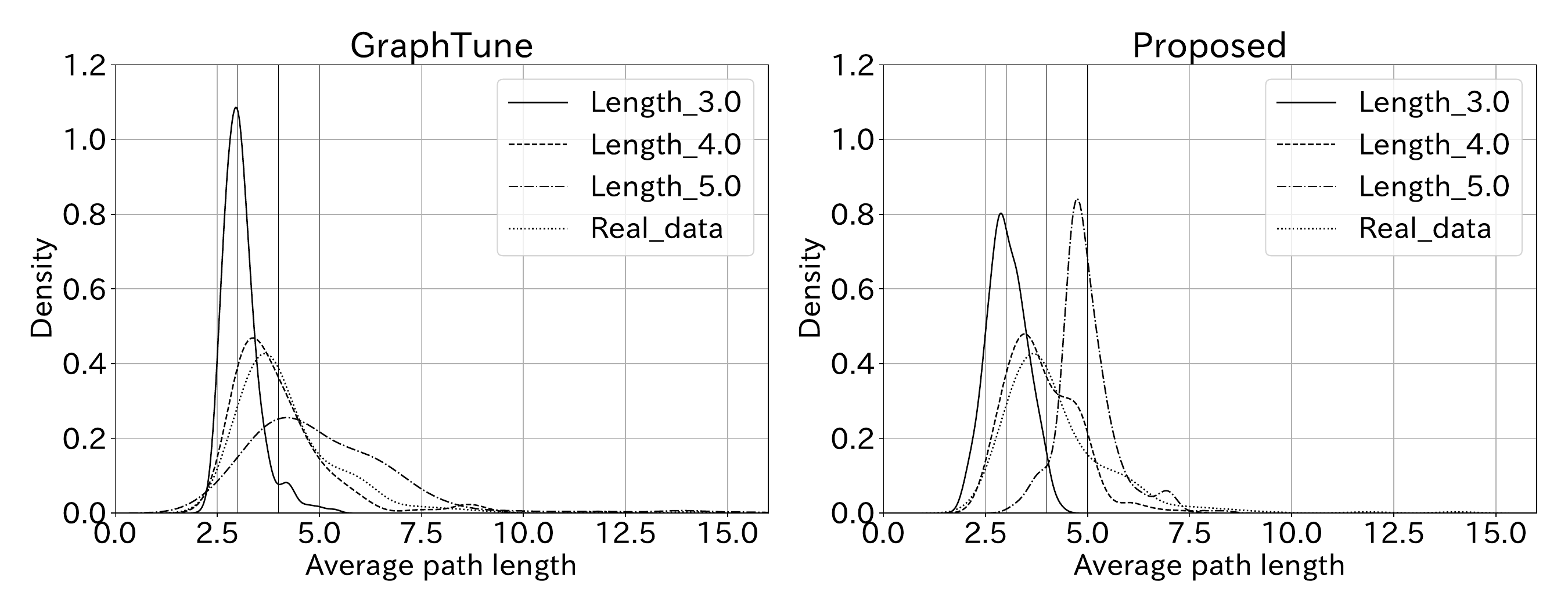}
  \vspace*{-1em}
  \caption{Kernel density estimation plots for the generated graphs when average shortest path length is specified as 3.0, 4.0, or 5.0.}
  \vspace*{-1em}
  \label{fig:generate_histogram}
\end{figure}

We propose an accurate generative model for graphs that extends GraphTune by adding an LSTM-based model called a feature estimator (see Fig.~\ref{fig:proposed_model}). 
The feature estimator estimates value of features of a generated graph and adds an error between the estimated values and elements of a condition vector (i.e. specified values of features) to the overall loss. 
In other words, the feature estimator feeds back information on features in generated graphs to neural networks in the GraphTune part, thereby allowing more accurate tuning of features. 
To appropriately train the proposed model, we also propose a training algorithm to avoid target leakage. 

\subsubsection*{GraphTune part}
  GraphTune part in our model learns to reconstruct a sequence from input sequence and a condition vector in the same manner as the original GraphTune.
  The details of the process follow the paper of GraphTune~\cite{b1}.
  In the generation step, only GraphTune part is used. 

\subsubsection*{Feature estimator}
  The feature estimator learns to estimate values of features of an output graph from a sequence reconstructed by GraphTune part.
  The reconstructed sequences are recursively input to the LSTM block. 
  The output of the last LSTM block is an estimator of values of features we focused on.
  The error between the estimator and the values of the features of the input graph is added to the loss of the GraphTune part to provide direct feedback on the accuracy of feature reproduction.
  Note that the feature estimator takes only reconstructed sequences as input and does not use the condition vector.


In the training of the proposed model, information of a condition vector (i.e. the value of the feature of input graphs) that is input to GraphTune part may leak to the feature estimator side, and appropriate training cannot be expected.
Therefore, an alternate training algorithm is applied to train the GraphTune part and the feature estimator alternately.
In the algorithm, when training one model, the parameters of the other model are freezed.
By repeating this alternate training, the dependency between the two models is eliminated and appropriate training is achieved.


\section{Experiments}

We train the models and generate graphs using a real graph dataset to verify the effectiveness of the proposed model in the accuracy of graph generation.
We use 2,000 induced subgraphs sampled from the Twitter who-follows-whom graph in the Higgs Twitter Dataset~\cite{b2} for training in order to compare the proposed model with GraphTune. 
The average shortest path length was used as the condition vector, and three patterns were specified at generation: 3.0, 4.0, and 5.0, each of which was set to generate 300 graphs.

  The parameters of the proposed model are set as follows.
  The size of the single fully connected layer placed before the feature estimator is 256, the size of the hidden state vector of the LSTM is 512, and the size of the last single fully connected layer is 1.
  The parameters of the GraphTune part in the proposed model and GraphTune as a comparison method were set according to the paper of GraphTune~\cite{b1}.
  The batch size was set to 37, and the number of epochs in feature estimator training was set to 10,000.
  The number of iterations in the alternate training was set to 2.

To evaluate the tuning accuracy of graph features, the distribution of the average shortest path length of the generated graphs is shown Fig.~\ref{fig:generate_histogram}.
Depending on the specified value of the average shortest path length, the two models show a change in distribution. 
Especially when 5.0 is specified, indicating that the graphs generated by the proposed model are distributed closer to the specified value than GraphTune.

\section{Conclusion}

We extended GraphTune, a conventional generative model with tunable features, by adding a feature estimator model that estimates values of features on generated graphs. 
We also proposed an alternate training algorithm to ensure that GraphTune and the feature estimator cooperate with each other and learn appropriately.
Through experiments with a real-world graph dataset, we confirmed that the proposed model can specify features more accurately than the original GraphTune.
We plan to apply various real-world graph datasets and specify different graph features to verify the generalization performance of the proposed model.

\bibliographystyle{IEEEtran}


\end{document}